# Deep Learning for Sentiment Analysis: A Survey

Lei Zhang, LinkedIn Corporation, lzhang32@gmail.com
Shuai Wang, University of Illinois at Chicago, shuaiwanghk@gmail.com
Bing Liu, University of Illinois at Chicago, liub@uic.edu

**Abstract**

Deep learning has emerged as a powerful machine learning technique that learns multiple layers of representations or features of the data and produces state-of-the-art prediction results. Along with the success of deep learning in many other application domains, deep learning is also popularly used in sentiment analysis in recent years. This paper first gives an overview of deep learning and then provides a comprehensive survey of its current applications in sentiment analysis.

## INTRODUCTION

Sentiment analysis or opinion mining is the computational study of people's opinions, sentiments, emotions, appraisals, and attitudes towards entities such as products, services, organizations, individuals, issues, events, topics, and their attributes.[1] The inception and rapid growth of the field coincide with those of the social media on the Web, for example, reviews, forum discussions, blogs, micro-blogs, Twitter, and social networks, because for the first time in human history, we have a huge volume of opinionated data recorded in digital forms. Since early 2000, sentiment analysis has grown to be one of the most active research areas in natural language processing (NLP). It is also widely studied in data mining, Web mining, text mining, and information retrieval. In fact, it has spread from computer science to management sciences and social sciences such as marketing, finance, political science, communications, health science, and even history, due to its importance to business and society as a whole. This proliferation is due to the fact that opinions are central to almost all human activities and are key influencers of our behaviours. Our beliefs and perceptions of reality, and the choices we make, are, to a considerable degree, conditioned upon how others see and evaluate the world. For this reason, whenever we need to make a decision we often seek out the opinions of others. This is not only true for individuals but also true for organizations.

Nowadays, if one wants to buy a consumer product, one is no longer limited to asking one's friends and family for opinions because there are many user reviews and discussions about the product in public forums on the Web. For an organization, it may no longer be necessary to conduct surveys, opinion polls, and focus groups in order to gather public opinions because there is an abundance of such information publicly available. In recent years, we have witnessed that opinionated postings in social media have helped reshape businesses, and sway public sentiments and emotions, which have profoundly impacted on our social and political systems. Such postings have also mobilized masses for political changes such as those happened in some Arab countries in 2011. It has thus become a necessity to collect and study opinions[1].

However, finding and monitoring opinion sites on the Web and distilling the information contained in them remains a formidable task because of the proliferation of diverse sites. Each site typically contains a huge volume of opinion text that is not always easily deciphered in long blogs and forum postings. The average human reader will have difficulty identifying relevant sites and extracting and summarizing the opinions in them. Automated sentiment analysis systems are thus needed. Because of this, there are many start-ups focusing on providing sentiment analysis services. Many big corporations have also built their own in-house capabilities. These practical applications and industrial interests have provided strong motivations for research in sentiment analysis.

Existing research has produced numerous techniques for various tasks of sentiment analysis, which include both supervised and unsupervised methods. In the supervised setting, early papers used all types of supervised machine learning methods (such as Support Vector Machines (SVM), Maximum Entropy, Naïve Bayes, etc.) and feature combinations. Unsupervised methods include various methods that exploit sentiment lexicons, grammatical analysis, and syntactic patterns. Several survey books and papers have been published, which cover those early methods and applications extensively.[1,2,3]

Since about a decade ago, deep learning has emerged as a powerful machine learning technique[4] and produced state-of-the-art results in many application domains, ranging from computer vision and speech recognition to NLP. Applying deep learning to sentiment analysis has also become very popular recently. This paper first gives an overview of deep learning and then provides a comprehensive survey of the sentiment analysis research based on deep learning.

**NEURAL NETWORKS**

Deep learning is the application of **artificial neural networks** (neural networks for short) to learning tasks using networks of multiple layers. It can exploit much more learning (representation) power of neural networks, which once were deemed to be practical only with one or two layers and a small amount of data.

Inspired by the structure of the biological brain, neural networks consist of a large number of information processing units (called neurons) organized in layers, which work in unison. It can learn to perform tasks (e.g., classification) by adjusting the connection weights between neurons, resembling the learning process of a biological brain.

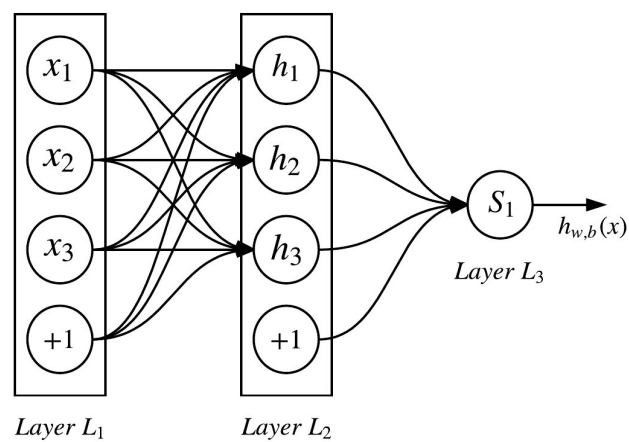

Figure 1: Feedforward neural network

Based on network topologies, neural networks can generally be categorized into feedforward neural networks and recurrent/recursive neural networks, which can also be mixed and matched. We will describe recurrent/recursive neural networks later. A simple example of a feedforward neural network is given in Figure 1, which consists of three layers $L_1$, $L_2$ and $L_3$. $L_1$ is the **input layer**, which corresponds to the input vector $(x_1, x_2, x_3)$ and intercept term +1. $L_3$ is the **output layer**, which corresponds to the output vector $(s_1)$. $L_2$ is the **hidden layer**, whose output is not visible as a network output. A circle in $L_1$ represents an element in the input vector, while a circle in $L_2$ or $L_3$ represents a neuron, the basic computation element of a neural network. We also call it an **activation function**. A line between two neurons represents a connection for the flow of information. Each connection is associated with a **weight**, a value controlling the signal between two neurons. The learning of a neural network is achieved by adjusting the weights between neurons with the information flowing through them. Neurons read output from neurons in the previous layer, process the information, and then generate output to neurons in the next layer. As in Figure 1, the neutral network alters weights based on training examples $(x^{(i)}, y^{(i)})$. After the training process, it will obtain a complex form of hypotheses $h_{w,b}(x)$ that fits the data.

Diving into the hidden layer, we can see that each neuron in $L_2$ takes input $x_1, x_2, x_3$ and intercept +1 from $L_1$, and outputs a value $f(W^t x) = f(\sum_{i=1}^{3} W_i x_i + b)$ by the activation function $f$. $W_i$ are weights of the connections; $b$ is the intercept or bias; $f$ is normally non-linear. The common choices of $f$ are **sigmoid** function, hyperbolic tangent function (**tanh**), or rectified linear function (**ReLU**). Their equations are as follows.

$$f(W^t x) = sigmoid(W^t x) = \frac{1}{1+\exp(-W^t x)} \quad (1)$$

$$f(W^t x) = \tanh(W^t x) = \frac{e^{W^t x} - e^{-W^t x}}{e^{W^t x} + e^{-W^t x}} \quad (2)$$

$$f(W^t x) = ReLU(W^t x) = \max(0, W^t x) \quad (3)$$

The sigmoid function takes a real-valued number and squashes it to a value in the range between 0 and 1. The function has been in frequent use historically due to its nice interpretation as the firing rate of a neuron: 0 for not firing or 1 for firing. But the non-linearity of the sigmoid has recently fallen out of favour because its activations can easily saturate at either tail of 0 or 1, where gradients are almost zero and the information flow would be cut. What is more is that its output is not zero-centered, which could introduce undesirable zig-zagging dynamics in the gradient updates for the connection weights in training. Thus, the tanh function is often more preferred in practice as its output range is zero-centered, [-1, 1] instead of [0, 1]. The ReLU function has also become popular lately. Its activation is simply thresholded at zero when the input is less than 0. Compared with the sigmoid function and the tanh function, ReLU is easy to compute, fast to converge in training and yields equal or better performance in neural networks.[5]

In $L_3$, we can use the **softmax** function as the output neuron, which is a generalization of the logistic function that squashes a *K*-dimensional vector $X$ of arbitrary real values to a *K*-dimensional vector $\sigma(X)$ of real values in the range (0, 1) that add up to 1. The function definition is as follows.

$$\sigma(X)_j = \frac{e^{x_j}}{\sum_{k=1}^{K} e^{x_k}} \quad for\ j = 1, \dots, k \quad (4)$$

Generally, softmax is used in the final layer of neural networks for final classification in feedforward neural networks.

By connecting together all neurons, the neural network in Figure 1 has parameters $(W, b) = (W^{(1)}, b^{(1)}, W^{(2)}, b^{(2)})$, where $W_{ij}^{(l)}$ denotes the weight associated with the connection between neuron $j$ in layer $l$, and neuron $i$ in layer $l + 1$. $b_i^{(l)}$ is the bias associated with neuron $i$ in layer $l + 1$.

To train a neural network, **stochastic gradient descent** via **backpropagation**[6] is usually employed to minimize the **cross-entropy** loss, which is a loss function for softmax output. Gradients of the loss function with respect to weights from the last hidden layer to the output layer are first calculated, and then gradients of the expressions with respect to weights between upper network layers are calculated recursively by applying the chain rule in a backward manner. With those gradients, the weights between layers are adjusted accordingly. It is an iterative refinement process until certain stopping criteria are met. The pseudo code for training the neural network in Figure 1 is as follows.

> **Training algorithm**: stochastic gradient descent via backpropagation
>
> Initialize weights $W$ and biases $b$ of the neural network $N$ with random values
> **do**
>    **for each** training example $(x_i, y_i)$
>       $p_i$ = neural-network-prediction $(N, x_i)$
>       calculate gradients of loss function $(p_i, y_i)$ with respect to $w^2$ at layer $L_3$
>       get $\Delta w^2$ for all weights from hidden layer $L_2$ to output layer $L_3$
>       calculate gradient with respect to $w^1$ by chain rule at layer $L_2$
>       get $\Delta w^1$ for all weights from input layer $L_1$ to hidden layer $L_2$
>       update ($w^1, w^2$)
> **until** all training examples are classified correctly or other stopping criteria are met
> **return** the trained neural network

Table 1: Training the neural network in Figure 1.

The above algorithm can be extended to generic feedforward neural network training with multiple hidden layers. Note that stochastic gradient descent estimates the parameters for every training example as opposed to the whole set of training examples in **batch gradient descent**. Therefore, the parameter updates have a high variance and cause the loss function to fluctuate to different intensities, which helps discover new and possibly better local minima.

**DEEP LEARNING**

The research community lost interests in neural networks in late 1990s mainly because they were regarded as only practical for "shallow" neural networks (neural networks with one or two layers) as training a "deep" neural network (neural networks with more layers) is complicated and computationally very expensive. However, in the past 10 years, deep learning made breakthrough and produced state-of-the-art results in many application domains, starting from computer vision, then speech recognition, and more recently, NLP.[7,8] The renaissance of neural networks can be attributed to many factors. Most important ones include: (1) the availability of computing power due to the advances in hardware (e.g., GPUs), (2) the availability of huge amounts of training data, and (3) the power and flexibility of learning intermediate representations.[9]

In a nutshell, deep learning uses a cascade of multiple layers of nonlinear processing units for feature extraction and transformation. The lower layers close to the data input learn simple features, while higher layers learn more complex features derived from lower layer features. The architecture forms a hierarchical and powerful feature representation. Figure 2 shows the feature hierarchy from the left (a lower layer) to the right (a higher layer) learned by deep learning in face image classification.[10] We can see that the learned image features grow in complexity, starting from blobs/edges, then noses/eyes/cheeks, to faces.

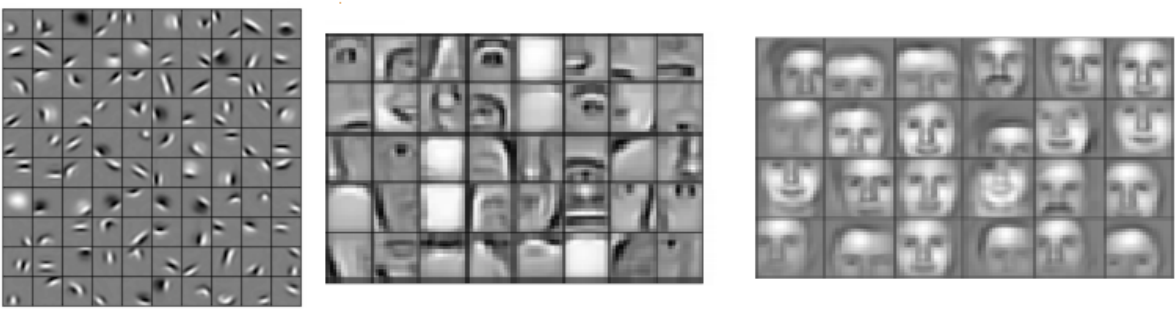

Figure 2: Feature hierarchy by deep learning

In recent years, deep learning models have been extensively applied in the field of NLP and show great potentials. In the following several sections, we briefly describe the main deep learning architectures and related techniques that have been applied to NLP tasks.

## WORD EMBEDDING

Many deep learning models in NLP need **word embedding** results as input features.[7] Word embedding is a technique for language modelling and feature learning, which transforms words in a vocabulary to vectors of continuous real numbers (e.g., $word\ "hat" \rightarrow (\ldots, 0.15, \ldots, 0.23, \ldots, 0.41, \ldots)$ ). The technique normally involves a mathematic embedding from a high-dimensional sparse vector space (e.g., one-hot encoding vector space, in which each word takes a dimension) to a lower-dimensional dense vector space. Each dimension of the embedding vector represents a latent feature of a word. The vectors may encode linguistic regularities and patterns.

The learning of word embeddings can be done using neural networks[11-15] or matrix factorization.[16,17] One commonly used word embedding system is **Word2Vec**[i]**,** which is essentially a computationally-efficient neural network prediction model that learns word embeddings from text. It contains **Continuous Bag-of-Words** model (CBOW)[13], and **Skip-Gram** model (SG)[14]. The CBOW model predicts the target word (e.g., "*wearing*") from its context words ("*the boy is _ a hat*", where "_" denotes the target word), while the SG model does the inverse, predicting the context words given the target word. Statistically, the CBOW model smoothens over a great deal of distributional information by treating the entire context as one observation. It is effective for smaller datasets. However, the SG model treats each context-target pair as a new observation and is better for larger datasets. Another frequently used learning approach is **Global Vector**[ii] (GloVe)[17], which is trained on the non-zero entries of a global word-word co-occurrence matrix.

## AUTOENCODER AND DENOISING AUTOENCODER

**Autoencoder Neural Network** is a three-layer neural network, which sets the target values to be equal to the input values. Figure 3 shows an example of an autoencoder architecture.

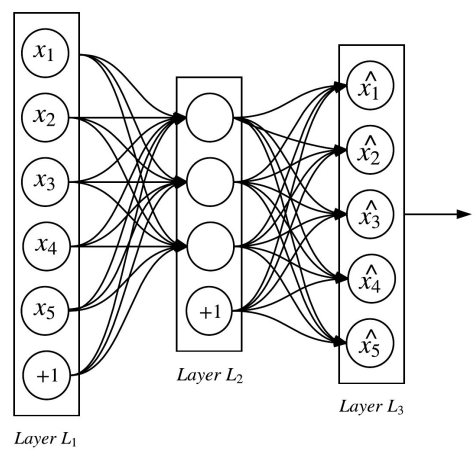

Figure 3: Autoencoder neural network

---

[i] Source code: https://code.google.com/archive/p/word2vec/
[ii] Source code: https://github.com/stanfordnlp/GloVe

Given the input vector $x \in [0,1]^d$, the autoencoder first maps it to a hidden representation $y \in [0,1]^{d'}$ by an encoder function $h(\cdot)$ (e.g., the sigmoid function). The latent representation $y$ is then mapped back by a decoder function $g(\cdot)$ into a reconstruction $r(x) = g(h(x))$. The autoencoder is typically trained to minimize a form of reconstruction error $loss(x, r(x))$. The objective of the autoencoder is to learn a representation of the input, which is the activation of the hidden layer. Due to the nonlinear function $h(\cdot)$ and $g(\cdot)$, the autoencoder is able to learn non-linear representations, which give it much more expressive power than its linear counterparts, such as **Principal Component Analysis** (PCA) or **Latent Semantic Analysis** (LSA).

One often stacks autoencoders into layers. A higher level autoencoder uses the output of the lower one as its training data. The stacked autoencoders[18] along with **Restricted Boltzmann Machines** (RBMs)[19] are earliest approaches to building deep neural networks. Once a stack of autoencoders has been trained in an **unsupervised** fashion, their parameters describing multiple levels of representations for $x$ (intermediate representations) can be used to initialize a supervised deep neural network, which has been shown empirically better than random parameter initialization.

The **Denoising Autoencoder** (DAE)[20] is an extension of autoencoder, in which the input vector $x$ is stochastically corrupted into a vector $\tilde{x}$. And the model is trained to denoise it, that is, to minimize a denoising reconstruction error $loss(x, r(\tilde{x}))$. The idea behind DAE is to force the hidden layer to discover more robust features and prevent it from simply learning the identity. A robust model should be able to reconstruct the input well even in the presence of noises. For example, deleting or adding a few of words from or to a document should not change the semantic of the document.

## CONVOLUTIONAL NEURAL NETWORK

**Convolutional Neural Network** (CNN) is a special type of feedforward neural network originally employed in the field of computer vision. Its design is inspired by the human visual cortex, a visual mechanism in animal brain. The visual cortex contains a lot of cells that are responsible for detecting light in small and overlapping sub-regions of the visual fields, which are called **receptive fields**. These cells act as local filters over the input space. CNN consists of multiple convolutional layers, each of which performs the function that is processed by the cells in the visual cortex.

Figure 4 shows a CNN for recognizing traffic signs.[21] The input is a 32x32x1 pixel image (32 x 32 represents image width x height; 1 represents input channel). In this first stage, the filter (size 5x5x1) is used to scan the image. Each region in the input image that the filter projects on is a receptive field. The filter is actually an array of numbers (called weights or parameters). As the filter is sliding (or **convolving**), it is multiplying its weight values with the original pixel values of the image (**element wise multiplications**). The multiplications are all summed up to a single number, which is a representative of the receptive field. Every receptive field produces a number. After the filter finishes scanning over the image, we can get an array (size 28x28x1), which is called the **activation map** or **feature map**. In CNN, we need to use different filters to scan the input. In Figure 4, we apply 108 kinds of filters and thus have 108 stacked feature maps in the first stage, which consists of the first convolutional layer. Following the convolutional layer, a **subsampling** (or **pooling**) layer is usually used to progressively reduce the spatial size of the representation, thus to reduce the number of features and the computational complexity of the network. For example, after subsampling in the first stage, the convolutional layer reduces its dimensions to (14x14x108). Note that while the dimensionality of each feature map is reduced, the subsampling step retains the most important information, with a commonly used subsampling operation being the max pooling. Afterwards, the output from the first stage becomes input to the second stage and the new filters are employed. The new filter size is 5x5x108, where 108 is the feature map size of the last layer. After the second stage, CNN uses a fully connected layer and then a softmax readout layer with output classes for classification.

Convolutional layers in CNN play the role of feature extractor, which extracts local features as they restrict the receptive fields of the hidden layers to be local. It means that CNN has a special spatially-local correlation by enforcing a local connectivity pattern between neurons of adjacent layers. Such a characteristic is useful for classification in NLP, in which we expect to find strong local clues regarding class membership, but these clues can appear in different places in the input. For example, in a document classification task, a single key phrase (or an n-gram) can help in determining the topic of the document. We would like to learn that certain sequences of words are good indicators of the topic, and do not necessarily care where they appear in the document. Convolutional and pooling layers allow the CNN to learn to find such local indicators, regardless of their positions.[8]

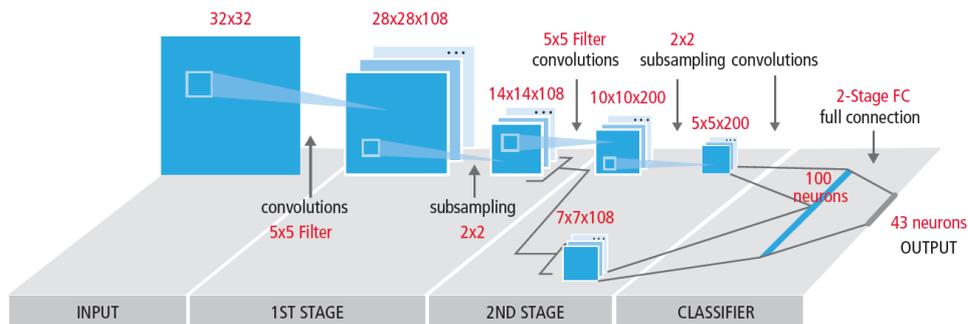

Figure 4: Convolutional neural network

## RECURRENT NEURAL NETWORK

**Recurrent Neural Network** (RNN)[22] is a class of neural networks whose connections between neurons form a directed cycle. Unlike feedforward neural networks, RNN can use its internal "memory" to process a sequence of inputs, which makes it popular for processing sequential information. The "memory" means that RNN performs the same task for every element of a sequence with each output being dependent on all previous computations, which is like "remembering" information about what has been processed so far.

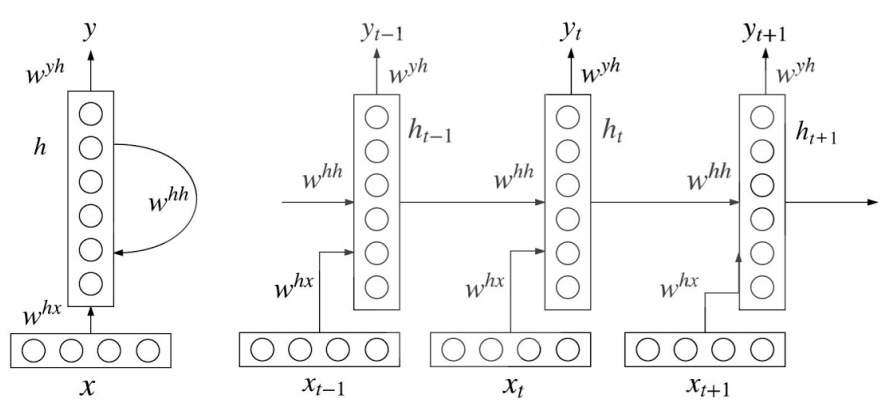

Figure 5: Recurrent neural network

Figure 5 shows an example of a RNN. The left graph is an unfolded network with cycles, while the right graph is a folded sequence network with three time steps. The length of time steps is determined by the length of input. For example, if the word sequence to be processed is a sentence of six words, the RNN would be unfolded into a neural network with six time steps or layers. One layer corresponds to a word.

In Figure 5, $x_t$ is the input vector at time step $t$. $h_t$ is the hidden state at time step $t$, which is calculated based on the previous hidden state and the input at the current time step.

$$h_t = f(w^{hh} h_{t-1} + w^{hx} x_t) \quad (5)$$

In Equation (5), the activation function $f$ is usually the tanh function or the ReLU function. $w^{hx}$ is the weight matrix used to condition the input $x_t$. $w^{hh}$ is the weight matrix used to condition the previous hidden state $h_{t-1}$.

$y_t$ is the output probability distribution over the vocabulary at step $t$. For example, if we want to predict the next word in a sentence, it would be a vector of probabilities across the word vocabulary.

$$y_t = softmax(w^{yh} h_t) \quad (6)$$

The hidden state $h_t$ is regarded as the memory of the network. It captures information about what happened in all previous time steps. $y_t$ is calculated solely based on the memory $h_t$ at time $t$ and the corresponding weight matrix $w^{yh}$.

Unlike a feedforward neural network, which uses different parameters at each layer, RNN shares the same parameters ($W^{hx}, W^{hh}, W^{yh}$) across all steps. This means that it performs the same task at each step, just with different inputs. This greatly reduces the total number of parameters needed to learn.

Theoretically, RNN can make use of the information in arbitrarily long sequences, but in practice, the standard RNN is limited to looking back only a few steps due to the **vanishing gradient** or **exploding gradient** problem.[23]

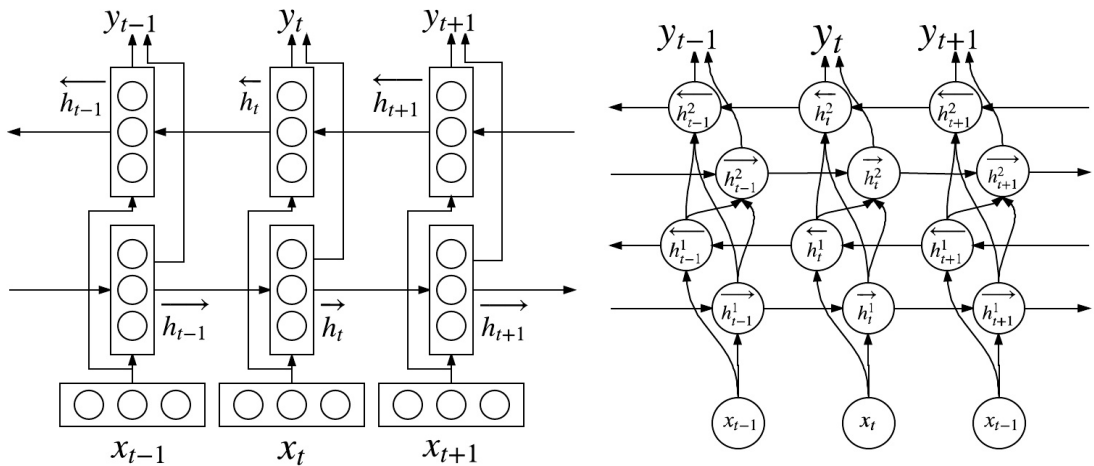

Figure 6: Bidirectional RNN (left) and deep bidirectional RNN (right)

Researchers have developed more sophisticated types of RNN to deal with the shortcomings of the standard RNN model: **Bidirectional RNN**, **Deep Bidirectional RNN** and **Long Short Term Memory** network. Bidirectional RNN is based on the idea that the output at each time may not only depend on the previous elements in the sequence, but also depend on the next elements in the sequence. For instance, to predict a missing word in a sequence, we may need to look at both the left and the right context. A bidirectional RNN[24] consists of two RNNs, which are stacked on the top of each other. The one that processes the input in its original order and the one that processes the reversed input sequence. The output is then computed based on the hidden state of both RNNs. Deep bidirectional RNN is similar to bidirectional RNN. The only difference is that it has multiple layers per time step, which provides higher learning capacity but needs a lot of training data. Figure 6 shows examples of bidirectional RNN and deep bidirectional RNN (with two layers) respectively.

## LSTM NETWORK

**Long Short Term Memory** network (LSTM)[25] is a special type of RNN, which is capable of learning long-term dependencies.

All RNNs have the form of a chain of repeating modules. In standard RNNs, this repeating module normally has a simple structure. However, the repeating module for LSTM is more complicated. Instead of having a single neural network layer, there are four layers interacting in a special way. Besides, it has two states: hidden state and cell state.

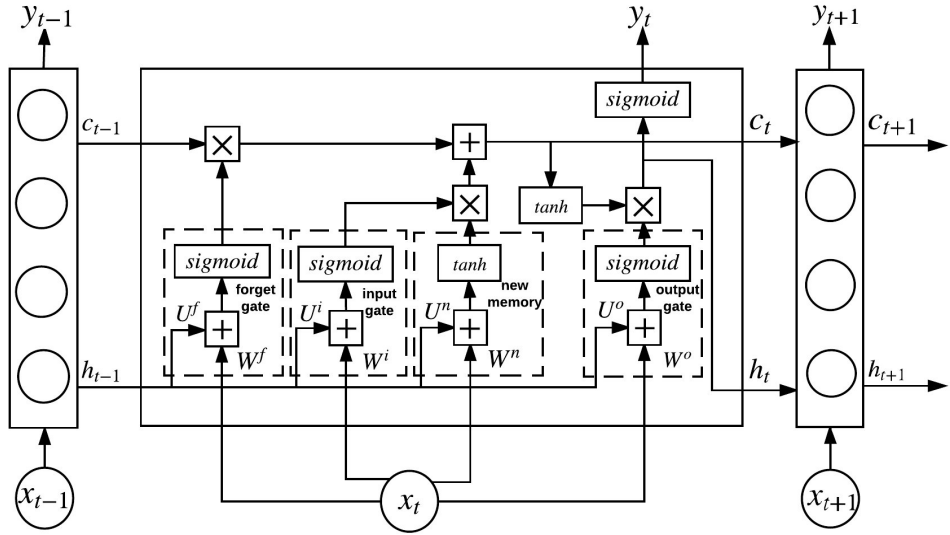

Figure 7: Long Short Term Memory network

Figure 7 shows an example of LSTM. At time step $t$, LSTM first decides what information to dump from the cell state. This decision is made by a sigmoid function/layer $\sigma$, called the "**forget gate**". The function takes $h_{t-1}$ (output from the previous hidden layer) and $x_t$ (current input), and outputs a number in [0, 1], where 1 means "completely keep" and 0 means "completely dump" in Equation (7).

$$f_t = \sigma(W^f x_t + U^f h_{t-1}) \qquad (7)$$

Then LSTM decides what new information to store in the cell state. This has two steps. First, a sigmoid function/layer, called the "**input gate**" as Equation (8), decides which values LSTM will update. Next, a tanh function/layer creates a vector of new candidate values $\widetilde{C}_t$, which will be added to the cell state. LSTM combines these two to create an update to the state.

$$i_t = \sigma(W^i x_t + U^i h_{t-1}) \qquad (8)$$

$$\widetilde{C}_t = \tanh(W^n x_t + U^n h_{t-1}) \qquad (9)$$

It is now time to update the old cell state $C_{t-1}$ into new cell state $C_t$ as Equation (10). Note that forget gate $f_t$ can control the gradient passes through it and allow for explicit "memory" deletes and updates, which helps alleviate vanishing gradient or exploding gradient problem in standard RNN.

$$C_t = f_t * C_{t-1} + i_t * \widetilde{C}_t \qquad (10)$$

Finally, LSTM decides the output, which is based on the cell state. LSTM first runs a sigmoid layer, which decides which parts of the cell state to output in Equation (11), called "**output gate**". Then, LSTM puts the cell state through the tanh function and multiplies it by the output of the sigmoid gate, so that LSTM only outputs the parts it decides to as Equation (12).

$$o_t = \sigma(W^o x_t + U^o h_{t-1}) \tag{11}$$

$$h_t = o_t * \tanh(C_t) \tag{12}$$

LSTM is commonly applied to sequential data but can also be used for tree-structured data. Tai et al.[26] introduced a generalization of the standard LSTM to **Tree-structured LSTM** (Tree-LSTM) and showed better performances for representing sentence meaning than a sequential LSTM.

A slight variation of LSTM is the **Gated Recurrent Unit** (GRU).[27,28] It combines the "forget" and "input" gates into a single update gate. It also merges the cell state and hidden state, and makes some other changes. The resulting model is simpler than the standard LSTM model, and has been growing in popularity.

**ATTENTION MECHANISM WITH RECURRENT NEURAL NETWORK**

Supposedly, bidirectional RNN and LSTM should be able to deal with long-range dependencies in data. But in practice, the long-range dependencies are still problematic to handle. Thus, a technique called the **Attention Mechanism** was proposed.

The attention mechanism in neural networks is inspired by the visual attention mechanism found in humans. That is, the human visual attention is able to focus on a certain region of an image with "high resolution" while perceiving the surrounding image in "low resolution" and then adjusting the focal point over time. In NLP, the attention mechanism allows the model to learn what to attend to based on the input text and what it has produced so far, rather than encoding the full source text into a fixed-length vector like standard RNN and LSTM.

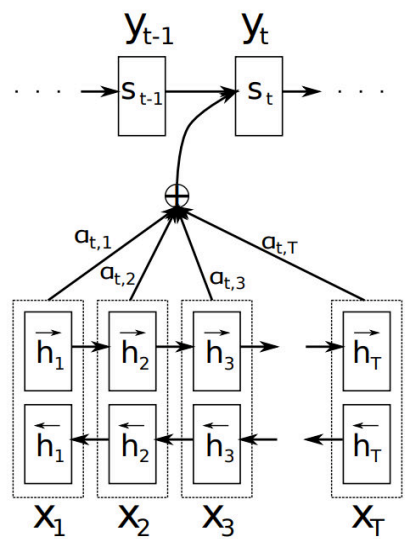

Figure 8: Attention mechanism in bidirectional recurrent neural network

Bahdanau et al.[29] first utilized the attention mechanism for machine translation in NLP. They proposed an encoder-decoder framework where an attention mechanism is used to select reference words in the original language for words in the target language before translation. Figure 8 illustrates the use of the attention mechanism in their bidirectional RNN. Note that each decoder output word $y_t$ depends on a weighted combination of all the input states, not just the last state as in the normal case. $a_{t,T}$ are weights that define in how much of each input state should be weighted for each output. For example, if $a_{2,2}$ has a big value, it means that the decoder pays a lot of **attention** to the second state in the source sentence while producing the second word of the target sentence. The weights of $a_{t,T}$ sum to 1 normally.

## MEMORY NETWORK

Weston et al.[30] introduced the concept of **Memory Networks** (MemNN) for the question answering problem. It works with several inference components combined with a large long-term memory. The components can be neural networks. The memory acts as a dynamic knowledge base. The four learnable/inference components function as follows: **I component** coverts the incoming input to the internal feature representation; **G component** updates old memories given the new input; **O component** generates output (also in the feature representation space); **R component** converts the output into a response format. For instance, given a list of sentences and a question for question answering, MemNN finds evidences from those sentences and generates an answer. During inference, the I component reads one sentence at a time and encodes it into a vector representation. Then the G component updates a piece of memory based on the current sentence representation. After all sentences are processed, a memory matrix (each row representing a sentence) is generated, which stores the semantics of the sentences. For a question, MemNN encodes it into a vector representation, then the O component uses the vector to select some related evidences from the memory and generates an output vector. Finally, the R component takes the output vector as the input and outputs a final response.

Based on MemNN, Sukhbaatar et al.[31] proposed an **End-to-End Memory Network** (MemN2N), which is a neural network architecture with a recurrent attention mechanism over the long-term memory component and it can be trained in an End-to-End manner through standard backpropagation. It demonstrates that multiple computational layers (hops) in the O component can uncover more abstractive evidences than a single layer and yield improved results for question answering and language modelling. It is worth noting that each computational layer can be a content-based attention model. Thus, MemN2N refines the attention mechanism to some extent. Note also a similar idea is the **Neural Turing Machines** reported by Graves et al.[32]

## RECURSIVE NEURAL NETWORK

**Recursive Neural Network** (RecNN) is a type of neural network that is usually used to learn a directed acyclic graph structure (e.g., tree structure) from data. A recursive neural network can be seen as a generalization of the recurrent neural network. Given the structural representation of a sentence (e.g., a parse tree), RecNN recursively generates parent representations in a bottom-up fashion, by combining tokens to produce representations for phrases, eventually the whole sentence. The sentence level representation can then be used to make a final classification (e.g., sentiment classification) for a given input sentence. An example process of vector composition in RecNN is shown in Figure 9 [33]. The vector of node "*very interesting*" is composed from the vectors of the node "*very*" and the node "*interesting*". Similarly, the node "*is very interesting*" is composed from the phrase node "*very interesting*" and the word node "*is*".

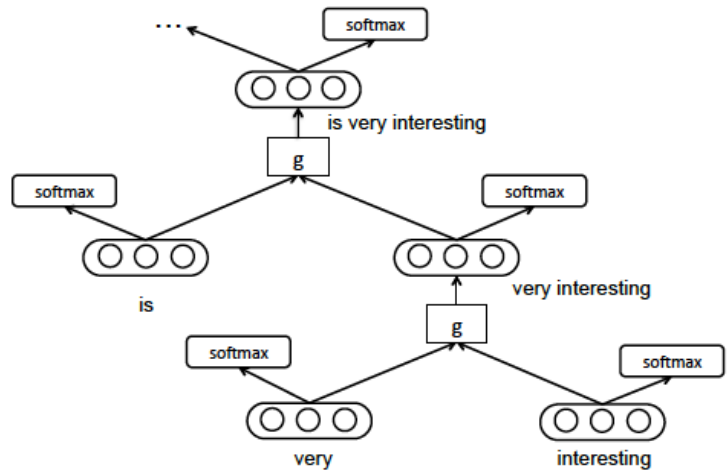

Figure 9: Recursive Neural network

**SENTIMENT ANALYSIS TASKS**

We are now ready to survey deep learning applications in sentiment analysis. But before doing that, we first briefly introduce the main sentiment analysis tasks in this section. For additional details, please refer to Liu's book[1] on sentiment analysis.

Researchers have mainly studied sentiment analysis at three levels of granularity: document level, sentence level, and aspect level. **Document level sentiment classification** classifies an opinionated document (e.g., a product review) as expressing an overall positive or negative opinion. It considers the whole document as the basic information unit and assumes that the document is known to be opinionated and contain opinions about a single entity (e.g., a particular phone). **Sentence level sentiment classification** classifies individual sentences in a document. However, each sentence cannot be assumed to be opinionated. Traditionally, one often first classifies a sentence as opinionated or not opinionated, which is called **subjectivity classification**. Then the resulting opinionated sentences are classified as expressing positive or negative opinions. Sentence level sentiment classification can also be formulated as a three-class classification problem, that is, to classify a sentence as neutral, positive or negative. Compared with document level and sentence level sentiment analysis, aspect level sentiment analysis or **aspect-based sentiment analysis** is more fine-grained. Its task is to extract and summarize people's opinions expressed on entities and aspects/features of entities, which are also called **targets**. For example, in a product review, it aims to summarize positive and negative opinions on different aspects of the product respectively, although the general sentiment on the product could be positive or negative. The whole task of aspect-based sentiment analysis consists of several subtasks such as **aspect extraction**, **entity extraction**, and **aspect sentiment classification**. For example, from the sentence, "*the voice quality of iPhone is great, but its battery sucks*", entity extraction should identify "*iPhone*" as the entity, and aspect extraction should identify that "*voice quality*" and "*battery*" are two aspects. Aspect sentiment classification should classify the sentiment expressed on the voice quality of the iPhone as positive and on the battery of the iPhone as negative. Note that for simplicity, in most algorithms aspect extraction and entity extraction are combined and are called aspect extraction or sentiment/opinion target extraction.

Apart from these core tasks, sentiment analysis also studies **emotion analysis, sarcasm detection**, **multilingual sentiment analysis**, etc. See Liu's book[1] for more details. In the following sections, we survey the deep learning applications in all these sentiment analysis tasks.

**DOCUMENT LEVEL SENTIMENT CLASSIFICATION**

Sentiment classification at the document level is to assign an overall sentiment orientation/polarity to an opinion document, i.e., to determine whether the document (e.g., a full online review) conveys an overall positive or negative opinion. In this setting, it is a binary classification task. It can also be formulated as a regression task, for example, to infer an overall rating score from 1 to 5 stars for the review. Some researchers also treat this as a 5-class classification task.

Sentiment classification is commonly regarded as a special case of document classification. In such a classification, document representation plays an important role, which should reflect the original information conveyed by words or sentences in a document. Traditionally, the bag-of-words model (BoW) is used to generate text representations in NLP and text mining, by which a document is regarded as a bag of its words. Based on BoW, a document is transformed to a numeric feature vector with a fixed length, each element of which can be the word occurrence (absence or presence), word frequency, or TF-IDF score. Its dimension equals to the size of the vocabulary. A document vector from BoW is normally very sparse since a single document only contains a small number of words in a vocabulary. Early neural networks adopted such feature settings.

Despite its popularity, BoW has some disadvantages. Firstly, the word order is ignored, which means that two documents can have exactly the same representation as long as they share the same words. Bag-of-N-Grams, an extension for BoW, can consider the word order in a short context (n-gram), but it also suffers from data sparsity and high dimensionality. Secondly, BoW can barely encode the semantics of words. For example, the words "*smart*", "*clever*" and "*book*" are of equal distance between them in BoW, but "*smart*" should be closer to "*clever*" than "*book*" semantically.

To tackle the shortcomings of BoW, word embedding techniques based on neural networks (introduced in the aforementioned section) were proposed to generate dense vectors (or low-dimensional vectors) for word representation, which are, to some extent, able to encode some semantic and syntactic properties of words. With word embeddings as input of words, document representation as a dense vector (or called dense document vector) can be derived using neural networks.

Notice that in addition to the above two approaches, i.e., using BoW and learning dense vectors for documents through word embeddings, one can also learn a dense document vector directly from BoW. We distinguish the different approaches used in related studies in Table 2.

When documents are properly represented, sentiment classification can be conducted using a variety of neural network models following the traditional supervised learning setting. In some cases, neural networks may only be used to extract text features/text representations, and these features are fed into some other non-neural classifiers (e.g., SVM) to obtain a final global optimum classifier. The properties of neural networks and SVM complement each other in such a way that their advantages are combined.

Besides sophisticated document/text representations, researchers also leveraged the characteristics of the data – product reviews, for sentiment classification. For product reviews, several researchers found it beneficial to jointly model sentiment and some additional information (e.g., user information and product information) for classification. Additionally, since a document often contains long dependency relations, the attention mechanism is also frequently used in document level sentiment classification. We summarize the existing techniques in Table 2.

| Research Work | Document/Text Representation | Neural Networks Model | Use Attention Mechanism | Joint Modelling with Sentiment |
|---|---|---|---|---|
| Moraes et al.[34] | BoW | ANN (Artificial Neural Network) | No | - |
| Le and Mikolov[35] | Learning dense vector at sentence, paragraph, document level | Paragraph Vector | No | - |
| Glorot et al.[36] | BoW to dense document vector | SDA (Stacked Denoising Autoencoder) | No | Unsupervised data representation from target domains (in transfer learning settings) |
| Zhai and Zhang[37] | BoW to dense document vector | DAE (Denoising Autoencoder) | No | - |
| Johnson and Zhang[38] | BoW to dense document vector | BoW-CNN and Seq-CNN | No | - |
| Tang et al.[39] | Word embeddings to dense document vector | CNN/LSTM (to learn sentence representation) + GRU (to learn document representation) | No | - |
| Tang et al.[40] | Word embeddings to dense document vector | UPNN (User Product Neutral Network) based on CNN | No | User information and product information |
| Chen et al.[41] | Word embeddings to dense document vector | UPA (User Product Attention) based on LSTM | Yes | User information and product Information |
| Dou[42] | Word embeddings to dense document vector | Memory Network | Yes | User information and product Information |
| Xu et al.[43] | Word embeddings to dense document vector | LSTM | No | - |
| Yang et al.[44] | Word embeddings to dense document vector | GRU-based sequence encoder | Hierarchical attention | - |
| Yin et al.[45] | Word embeddings to dense document vector | Input encoder and LSTM | Hierarchical attention | Aspect/target information |
| Zhou et al.[46] | Word embeddings to dense document vector | LSTM | Hierarchical attention | Cross-lingual information |
| Li et al.[47] | Word embeddings to dense document vector | Memory Network | Yes | Cross-domain information |

Table 2: Deep learning methods for document level sentiment classification

Below, we also give a brief description of these existing representative works.

Moraes et al.[34] made an empirical comparison between Support Vector Machines (SVM) and Artificial Neural Networks (ANN) for document level sentiment classification, which demonstrated that ANN produced competitive results to SVM's in most cases.

To overcome the weakness of BoW, Le and Mikolov[35] proposed Paragraph Vector, an unsupervised learning algorithm that learns vector representations for variable-length texts such as sentences, paragraphs and documents. The vector representations are learned by predicting the surrounding words in contexts sampled from the paragraph.

Glorot et al.[36] studied domain adaptation problem for sentiment classification. They proposed a deep learning system based on Stacked Denoising Autoencoder with sparse rectifier units, which can perform an unsupervised text feature/representation extraction using both labeled and unlabeled data. The features are highly beneficial for domain adaption of sentiment classifiers.

Zhai and Zhang[37] introduced a semi-supervised autoencoder, which further considers the sentiment information in its learning stage in order to obtain better document vectors, for sentiment classification. More specifically, the model learns a task-specific representation of the textual data by relaxing the loss function in the autoencoder to the Bregman Divergence and also deriving a discriminative loss function from the label information.

Johnson and Zhang[38] proposed a CNN variant named BoW-CNN that employs bag-of-word conversion in the convolution layer. They also designed a new model, called Seq-CNN, which keeps the sequential information of words by concatenating the one-hot vector of multiple words.

Tang et al.[39] proposed a neural network to learn document representation, with the consideration of sentence relationships. It first learns the sentence representation with CNN or LSTM from word embeddings. Then a GRU is utilized to adaptively encode semantics of sentences and their inherent relations in document representations for sentiment classification.

Tang et al.[40] applied user representations and product representations in review classification. The idea is that those representations can capture important global clues such as individual preferences of users and overall qualities of products, which can provide better text representations.

Chen et al.[41] also incorporated user information and product information for classification but via word and sentence level attentions, which can take into account of the global user preference and product characteristics at both the word level and the semantic level. Likewise, Dou[42] used a deep memory network to capture user and product information. The proposed model can be divided into two separate parts. In the first part, LSTM is applied to learn a document representation. In the second part, a deep memory network consisting of multiple computational layers (hops) is used to predict the review rating for each document.

Xu et al.[43] proposed a cached LSTM model to capture the overall semantic information in a long text. The memory in the model is divided into several groups with different forgetting rates. The intuition is to enable the memory groups with low forgetting rates to capture global semantic features and the ones with high forgetting rates to learn local semantic features.

Yang et al.[44] proposed a hierarchical attention network for document level sentiment rating prediction of reviews. The model includes two levels of attention mechanisms: one at the word level and the other at the sentence level, which allow the model to pay more or less attention to individual words or sentences in constructing the representation of a document.

Yin et al.[45] formulated the document-level aspect-sentiment rating prediction task as a machine comprehension problem and proposed a hierarchical interactive attention-based model. Specifically, documents and pseudo aspect-questions are interleaved to learn aspect-aware document representation.

Zhou et al.[46] designed an attention-based LSTM network for cross-lingual sentiment classification at the document level. The model consists of two attention-based LSTMs for bilingual representation, and each LSTM is also hierarchically structured. In this setting, it effectively adapts the sentiment information from a resource-rich language (English) to a resource-poor language (Chinese) and helps improve the sentiment classification performance.

Li et al.[47] proposed an adversarial memory network for cross-domain sentiment classification in a transfer learning setting, where the data from the source and the target domain are modelled together. It jointly trains two networks for sentiment classification and domain classification (i.e., whether a document is from the source or target domain).

## SENTENCE LEVEL SENTIMENT CLASSIFICATION

Sentence level sentiment classification is to determine the sentiment expressed in a single given sentence. As discussed earlier, the sentiment of a sentence can be inferred with subjectivity classification[48] and polarity classification, where the former classifies whether a sentence is subjective or objective and the latter decides whether a subjective sentence expresses a negative or positive sentiment. In existing deep learning models, sentence sentiment classification is usually formulated as a joint three-way classification problem, namely, to predict a sentence as positive, neural, and negative.

Same as document level sentiment classification, sentence representation produced by neural networks is also important for sentence level sentiment classification. Additionally, since a sentence is usually short compared to a document, some syntactic and semantic information (e.g., parse trees, opinion lexicons, and part-of-speech tags) may be used to help. Additional information such as review ratings, social relationship, and cross-domain information can be considered too. For example, social relationships have been exploited in discovering sentiments in social media data such as tweets.

In early research, parse trees (which provide some semantic and syntactic information) were used together with the original words as the input to neural models, so that the sentiment composition can be better inferred. But lately, CNN and RNN become more popular, and they do not need parse trees to extract features from sentences. Instead, CNN and RNN use word embeddings as input, which already encode some semantic and syntactic information. Moreover, the model architecture of CNN or RNN can help learn intrinsic relationships between words in a sentence too. The related works are introduced in detail below.

Socher et al.[49] first proposed a semi-supervised **Recursive Autoencoders Network** (RAE) for sentence level sentiment classification, which obtains a reduced dimensional vector representation for a sentence. Later on, Socher et al.[50] proposed a **Matrix-vector Recursive Neural Network** (MV-RNN), in which each word is additionally associated with a matrix representation (besides a vector representation) in a tree structure. The tree structure is obtained from an external parser. In Socher et al.[51], the authors further introduced the **Recursive Neural Tensor Network** (RNTN), where tensor-based compositional functions are used to better capture the interactions between elements. Qian et al.[33] proposed two more advanced models, **Tag-guided Recursive Neural Network** (TG-RNN), which chooses a composition function according to the part-of-speech tags of a phrase, and **Tag-embedded Recursive Neural Network / Recursive Neural Tenser Network** (TE-RNN/RNTN), which learns tag embeddings and then combines tag and word embeddings together.

Kalchbrenner et al.[52] proposed a **Dynamic CNN** (called DCNN) for semantic modelling of sentences. DCNN uses the dynamic K-Max pooling operator as a non-linear subsampling function. The feature graph induced by the network is able to capture word relations. Kim[53] also proposed to use CNN for sentence-level sentiment classification and experimented with several variants, namely CNN-rand (where word embeddings are randomly initialized), CNN-static (where word embeddings are pre-trained and fixed), CNN-non-static (where word embeddings are pre-trained and fine-tuned) and CNN-multichannel (where multiple sets of word embeddings are used).

dos Santos and Gatti[54] proposed a **Character to Sentence CNN** (CharSCNN) model. CharSCNN uses two convolutional layers to extract relevant features from words and sentences of any size to perform sentiment analysis of short texts. Wang et al.[55] utilized LSTM for Twitter sentiment classification by simulating the interactions of words during the compositional process. Multiplicative operations between word embeddings through gate structures are used to provide more flexibility and to produce better compositional results compared to the additive ones in simple recurrent neural network. Similar to bidirectional RNN, the unidirectional LSTM can be extended to a bidirectional LSTM[56] by allowing bidirectional connections in the hidden layer.

Wang et al.[57] proposed a regional CNN-LSTM model, which consists of two parts: regional CNN and LSTM, to predict the valence arousal ratings of text.

Wang et al.[58] described a joint CNN and RNN architecture for sentiment classification of short texts, which takes advantage of the coarse-grained local features generated by CNN and long-distance dependencies learned via RNN.

Guggilla et al.[59] presented a LSTM- and CNN-based deep neural network model, which utilizes word2vec and linguistic embeddings for claim classification (classifying sentences to be factual or feeling).

Huang et al.[60] proposed to encode the syntactic knowledge (e.g., part-of-speech tags) in a tree-structured LSTM to enhance phrase and sentence representation.

Akhtar et al.[61] proposed several multi-layer perceptron based ensemble models for fine-gained sentiment classification of financial microblogs and news.

Guan et al.[62] employed a weakly-supervised CNN for sentence (and also aspect) level sentiment classification. It contains a two-step learning process: it first learns a sentence representation weakly supervised by overall review ratings and then uses the sentence (and aspect) level labels for fine-tuning.

Teng et al.[63] proposed a context-sensitive lexicon-based method for sentiment classification based on a simple weighted-sum model, using bidirectional LSTM to learn the sentiment strength, intensification and negation of lexicon sentiments in composing the sentiment value of a sentence.

Yu and Jiang[64] studied the problem of learning generalized sentence embeddings for cross-domain sentence sentiment classification and designed a neural network model containing two separated CNNs that jointly learn two hidden feature representations from both the labeled and unlabeled data.

Zhao et al.[65] introduced a recurrent random walk network learning approach for sentiment classification of opinionated tweets by exploiting the deep semantic representation of both user posted tweets and their social relationships.

Mishra et al.[66] utilized CNN to automatically extract cognitive features from the eye-movement (or gaze) data of human readers reading the text and used them as enriched features along with textual features for sentiment classification.

Qian et al.[67] presented a linguistically regularized LSTM for the task. The proposed model incorporates linguistic resources such as sentiment lexicon, negation words and intensity words into the LSTM so as to capture the sentiment effect in sentences more accurately.

**ASPECT LEVEL SENTIMENT CLASSIFICATION**

Different from the document level and the sentence level sentiment classification, aspect level sentiment classification considers both the sentiment and the target information, as a sentiment always has a target. As mentioned earlier, a target is usually an entity or an entity aspect. For simplicity, both entity and aspect are usually just called aspect. Given a sentence and a target aspect, aspect level sentiment classification aims to infer the sentiment polarity/orientation of the sentence toward the target aspect. For example, in the sentence "*the screen is very clear but the battery life is too short.*" the sentiment is positive if the target aspect is "*screen*" but negative if the target aspect is "*battery life*". We will discuss automated aspect or target extraction in the next section.

Aspect level sentiment classification is challenging because modelling the semantic relatedness of a target with its surrounding context words is difficult. Different context words have different influences on the sentiment polarity of a sentence towards the target. Therefore, it is necessary capture semantic connections between the target word and the context words when building learning models using neural networks.

There are three important tasks in aspect level sentiment classification using neural networks. The first task is to represent the context of a target, where the context means the contextual words in a sentence or document. This issue can be similarly addressed using the text representation approaches mentioned in the above two sections. The second task is to generate a target representation, which can properly interact with its context. A general solution is to learn a target embedding, which is similar to word embedding. The third task is to identify the important sentiment context (words) for the specified target. For example, in the sentence "*the screen of iPhone is clear but batter life is short*", "*clear*" is the important context word for "*screen*" and "*short*" is the important context for "*battery life*". This task is recently addressed by the attention mechanism. Although many deep learning techniques have been proposed to deal with aspect level sentiment classification, to our knowledge, there are still no dominating techniques in the literature. Related works and their main focuses are introduced below.

Dong et al.[68] proposed an **Adaptive Recursive Neural Network** (AdaRNN) for target-dependent twitter sentiment classification, which learns to propagate the sentiments of words towards the target depending on the context and syntactic structure. It uses the representation of the root node as the features, and feeds them into the softmax classifier to predict the distribution over classes.

Vo and Zhang[69] studied aspect-based Twitter sentiment classification by making use of rich automatic features, which are additional features obtained using unsupervised learning methods. The paper showed that multiple embeddings, multiple pooling functions, and sentiment lexicons can offer rich sources of feature information and help achieve performance gains.

Since LSTM can capture semantic relations between the target and its context words in a more flexible way, Tang et al.[70] proposed **Target-dependent LSTM** (TD-LSTM) and **Target-connection LSTM** (TC-LSTM) to extend LSTM by taking the target into consideration. They regarded the given target as a feature and concatenated it with the context features for aspect sentiment classification.

Ruder et al.[71] proposed to use a hierarchical and bidirectional LSTM model for aspect level sentiment classification, which is able to leverage both intra- and inter-sentence relations. The sole dependence on sentences and their structures within a review renders the proposed model language-independent. Word embeddings are fed into a sentence-level bidirectional LSTM. Final states of the forward and backward LSTM are concatenated together with the target embedding and fed into a bidirectional review-level LSTM. At every time step, the output of the forward and backward LSTM is concatenated and fed into a final layer, which outputs a probability distribution over sentiments.

Considering the limitation of work by Dong et al.[68] and Vo and Zhang[69], Zhang et al.[72] proposed a sentence level neural model to address the weakness of pooling functions, which do not explicitly model tweet-level semantics. To achieve that, two gated neural networks are presented. First, a bi-directional gated neural network is used to connect the words in a tweet so that pooling functions can be applied over the hidden layer instead of words for better representing the target and its contexts. Second, a three-way gated neural network structure is used to model the interaction between the target mention and its surrounding contexts, addressing the limitations by using gated neural network structures to model the syntax and semantics of the enclosing tweet, and the interaction between the surrounding contexts and the target respectively. Gated neural networks have been shown to reduce the bias of standard recurrent neural networks towards the ends of a sequence by better propagation of gradients.

Wang et al.[73] proposed an attention-based LSTM method with target embedding, which was proven to be an effective way to enforce the neural model to attend to the related part of a sentence. The attention mechanism is used to enforce the model to attend to the important part of a sentence, in response to a specific aspect. Likewise, Yang et al.[74] proposed two attention-based bidirectional LSTMs to improve the classification performance. Liu and Zhang[75] extended the attention modelling by differentiating the attention obtained from the left context and the right context of a given target/aspect. They further controlled their attention contribution by adding multiple gates.

Tang et al.[76] introduced an end-to-end memory network for aspect level sentiment classification, which employs an attention mechanism with an external memory to capture the importance of each context word with respect to the given target aspect. This approach explicitly captures the importance of each context word when inferring the sentiment polarity of the aspect. Such importance degree and text representation are calculated with multiple computational layers, each of which is a neural attention model over an external memory.

Lei et al.[77] proposed to use a neural network approach to extract pieces of input text as rationales (reasons) for review ratings. The model consists of a generator and a decoder. The generator specifies a distribution over possible rationales (extracted text) and the encoder maps any such text to a task-specific target vector. For multi-aspect sentiment analysis, each coordinate of the target vector represents the response or rating pertaining to the associated aspect.

Li et al.[78] integrated the target identification task into sentiment classification task to better model aspect-sentiment interaction. They showed that sentiment identification can be solved with an end-to-end machine learning architecture, in which the two sub-tasks are interleaved by a deep memory network. In this way, signals produced in target detection provide clues for polarity classification, and reversely, the predicted polarity provides feedback to the identification of targets.

Ma et al.[79] proposed an **Interactive Attention Network** (IAN) that considers both attentions on target and context. That is, it uses two attention networks to interactively detect the important words of the target expression/description and the important words of its full context.

Chen et al.[80] proposed to utilize a recurrent attention network to better capture the sentiment of complicated contexts. To achieve that, their proposed model uses a recurrent/dynamic attention structure and learns non-linear combination of the attention in GRUs.

Tay et al.[81] designed a **Dyadic Memory Network** (DyMemNN) that models dyadic interactions between aspect and context, by using either neural tensor compositions or holographic compositions for memory selection operation.

## ASPECT EXTRACTION AND CATEGORIZATION

To perform aspect level sentiment classification, one needs to have aspects (or targets), which can be manually given or automatically extracted. In this section, we discuss existing work for automated aspect extraction (or aspect term extraction) from a sentence or document using deep learning models. Let us use an example to state the problem. For example, in the sentence *"the image is very clear"* the word *"image"* is an aspect term (or sentiment target). The associated problem of aspect categorization is to group the same aspect expressions into a category. For instance, the aspect terms *"image"*, *"photo"* and *"picture"* can be grouped into one aspect category named *Image*. In the review below, we include the extraction of both aspect and entity that are associated with opinions.

One reason why deep learning models can be helpful for this task is that, deep learning is essentially good at learning (complicated) feature representations. When an aspect is properly characterized in some feature space, for example, in one or some hidden layer(s), the semantics or correlation between an aspect and its context can be captured with the interplay between their corresponding feature representations. In other words, deep learning provides a possible approach to automated feature engineering without human involvement.

Katiyar and Cardie[82] investigated the use of deep bidirectional LSTMs for joint extraction of opinion entities and the IS-FROM and IS-ABOUT relationships that connect the entities. Wang et al.[83] further proposed a joint model integrating RNN and **Conditional Random Fields** (CRF) to co-extract aspects and opinion terms or expressions. The proposed model can learn high-level discriminative features and double-propagate information between aspect and opinion terms simultaneously. Wang et al.[84] further proposed a **Coupled Multi-Layer Attention Model** (CMLA) for co-extracting of aspect and opinion terms. The model consists of an aspect attention and an opinion attention using GRU units. An improved LSTM-based approach was reported by Li and Lam[85], specifically for aspect term extraction. It consists of three LSTMs, of which two LSTMs are for capturing aspect and sentiment interactions. The third LSTM is to use the sentiment polarity information as an additional guidance.

He et al.[86] proposed an attention-based model for unsupervised aspect extraction. The main intuition is to utilize the attention mechanism to focus more on aspect-related words while de-emphasizing aspect-irrelevant words during the learning of aspect embeddings, similar to the autoencoder framework.

Zhang et al.[87] extended a CRF model using a neural network to jointly extract aspects and corresponding sentiments. The proposed CRF variant replaces the original discrete features in CRF with continuous word embeddings, and adds a neural layer between the input and output nodes.

Zhou et al.[88] proposed a semi-supervised word embedding learning method to obtain continuous word representations on a large set of reviews with noisy labels. With the word vectors learned, deeper and hybrid features are learned by stacking on the word vectors through a neural network. Finally, a logistic regression classifier trained with the hybrid features is used to predict the aspect category.

Yin et al.[89] first learned word embedding by considering the dependency path connecting words. Then they designed some embedding features that consider the linear context and dependency context information for CRF-based aspect term extraction.

Xiong et al.[90] proposed an attention-based deep distance metric learning model to group aspect phrases. The attention-based model is to learn feature representation of contexts. Both aspect phrase embedding and context embedding are used to learn a deep feature subspace metric for K-means clustering.

Poria et al.[91] proposed to use CNN for aspect extraction. They developed a seven-layer deep convolutional neural network to tag each word in opinionated sentences as either aspect or non-aspect word. Some linguistic patterns are also integrated into the model for further improvement.

Ying et al.[92] proposed two RNN-based models for cross-domain aspect extraction. They first used rule-based methods to generate an auxiliary label sequence for each sentence. They then trained the models using both the true labels and auxiliary labels, which shows promising results.

**OPINION EXPRESSION EXTRACTION**

In this and the next few sections, we discuss deep learning applications to some other sentiment analysis related tasks. This section focuses on the problem of opinion expression extraction (or opinion term extraction, or opinion identification), which aims to identify the expressions of sentiment in a sentence or a document.

Similar to the aspect extraction, opinion expression extraction using deep learning models is workable because their characteristics could be identified in some feature space as well.

Irsoy and Cardie[93] explored the application of deep bidirectional RNN for the task, which outperforms traditional shallow RNNs with the same number of parameters and also previous CRF methods.[94]

Liu et al.[95] presented a general class of discriminative models based on the RNN architecture and word embedding. The authors used pre-trained word embeddings from three external sources in different RNN architectures including Elman-type, Jordan-type, LSTM and their variations.

Wang et al.[83] proposed a model integrating recursive neural networks and CRF to co-extract aspect and opinion terms. The aforementioned CMLA is also proposed for co-extraction of aspect and opinion terms.[84]

**SENTIMENT COMPOSITION**

Sentiment composition claims that the sentiment orientation of an opinion expression is determined by the meaning of its constituents as well as the grammatical structure. Due to their particular tree-structure design, RecNN is naturally suitable for this task.[51] Irsoy and Cardie[96] reported that the RecNN with a deep architecture can more accurately capture different aspects of compositionality in language, which benefits sentiment compositionality. Zhu et al.[97] proposed a neural network for integrating the compositional and non-compositional sentiment in the process of sentiment composition.

**OPINION HOLDER EXTRACTION**

Opinion holder (or source) extraction is the task of recognizing who holds the opinion (or whom/where the opinion is from).[1] For example, in the sentence "*John hates his car*", the opinion holder is "*john*". This problem is commonly formulated as a sequence labelling problem like opinion expression extraction or aspect extraction. Notice that opinion holder can be either explicit (from a noun phrase in the sentence) or implicit (from the writer) as shown by Yang and Cardie[98]. Deng and Wiebe[99] proposed to use word embeddings of opinion expressions as features for recognizing sources of participant opinions and non-participant opinions, where a source can be the noun phrase or writer.

**TEMPORAL OPINION MINING**

Time is also an important dimension in problem definition of sentiments analysis (see Liu's book[1]). As time passes by, people may maintain or change their mind, or even give new viewpoints.

Therefore, predicting future opinion is important in sentiment analysis. Some research using neural networks has been reported recently to tackle this problem.

Chen et al.[100] proposed a **Content-based Social Influence Model** (CIM) to make opinion behaviour predictions of twitter users. That is, it uses the past tweets to predict users' future opinions. It is based on a neural network framework to encode both the user content and social relation factor (one's opinion about a target is influenced by one's friends).

Rashkin et al.[101] used LSTMs for targeted sentiment forecast in the social media context. They introduced multilingual connotation frames, which aim at forecasting implied sentiments among world event participants engaged in a frame.

**SENTIMENT ANALYSIS WITH WORD EMBEDDING**

It is clear that word embeddings played an important role in deep learning based sentiment analysis models. It is also shown that even without the use of deep learning models, word embeddings can be used as features for non-neural learning models for various tasks. The section thus specifically highlights word embeddings' contribution to sentiment analysis.

We first present the works of sentiment-encoded word embeddings. For sentiment analysis, directly applying regular word methods like CBOW or Skip-gram to learn word embeddings from context can encounter problems, because words with similar contexts but opposite sentiment polarities (e.g., "*good*" or "*bad*") may be mapped to nearby vectors in the embedding space. Therefore, sentiment-encoded word embedding methods have been proposed. Mass el al.[102] learned word embeddings that can capture both semantic and sentiment information. Bespalov et al.[103] showed that an n-gram model combined with latent representation would produce a more suitable embedding for sentiment classification. Labutov and Lipson[104] re-embed existing word embeddings with logistic regression by regarding sentiment supervision of sentences as a regularization term.

Le and Mikolov[35] proposed the concept of paragraph vector to first learn fixed-length representation for variable-length pieces of texts, including sentences, paragraphs and documents. They experimented on both sentence and document-level sentiment classification tasks and achieved performance gains, which demonstrates the merit of paragraph vectors in capturing semantics to help sentiment classification. Tang et al.[105,106] presented models to learn **Sentiment-specific Word Embeddings** (SSWE), in which not only the semantic but also sentiment information is embedded in the learned word vectors. Wang and Xia[107] developed a neural architecture to train a sentiment-bearing word embedding by integrating the sentiment supervision at both the document and word levels. Yu et al.[108] adopted a refinement strategy to obtain joint semantic-sentiment bearing word vectors.

Feature enrichment and multi-sense word embeddings are also investigated for sentiment analysis. Vo and Zhang[69] studied aspect-based Twitter sentiment classification by making use of rich automatic features, which are additional features obtained using unsupervised learning techniques. Li and Jurafsky[109] experimented with the utilization of multi-sense word embeddings on various NLP tasks. Experimental results show that while such embeddings do improve the performance of some tasks, they offer little help to sentiment classification tasks. Ren et al.[110] proposed methods to learn topic-enriched multi-prototype word embeddings for Twitter sentiment classification.

Multilinguistic word embeddings have also been applied to sentiment analysis. Zhou et al.[111] reported a **Bilingual Sentiment Word Embedding** (BSWE) model for cross-language sentiment classification. It incorporates the sentiment information into English-Chinese bilingual embeddings by employing labeled corpora and their translation, instead of large-scale parallel corpora. Barnes et al.[112] compared several types of bilingual word embeddings and neural machine translation techniques for cross-lingual aspect-based sentiment classification.

Zhang et al.[113] integrated word embeddings with matrix factorization for personalized review-based rating prediction. Specifically, the authors refine existing semantics-oriented word vectors (e.g., word2vec and GloVe) using sentiment lexicons. Sharma et al.[114] proposed a semi-supervised technique to use sentiment bearing word embeddings for ranking sentiment intensity of adjectives. Word embedding techniques have also been utilized or improved to help address various sentiment analysis tasks in many other recent studies.[55,62,87,89,95]

## SARCASM ANALYSIS

Sarcasm is a form verbal irony and a closely related concept to sentiment analysis. Recently, there is a growing interest in NLP communities in sarcasm detection. Researchers have attempted to solve it using deep learning techniques due of their impressive success in many other NLP problems.

Zhang et al.[115] constructed a deep neural network model for tweet sarcasm detection. Their network first uses a bidirectional GRU model to capture the syntactic and semantic information over tweets locally, and then uses a pooling neural network to extract contextual features automatically from history tweets for detecting sarcastic tweets.

Joshi et al.[116] investigated word embeddings-based features for sarcasm detection. They experimented four past algorithms for sarcasm detection with augmented word embeddings features and showed promising results.

Poria et al.[117] developed a CNN-based model for sarcasm detection (sarcastic or non-sarcastic tweets classification), by jointly modelling pre-trained emotion, sentiment and personality features, along with the textual information in a tweet.

Peled and Reichart[118] proposed to interpret sarcasm tweets based on a RNN neural machine translation model.

Ghosh and Veale[119] proposed a CNN and bidirectional LSTM hybrid for sarcasm detection in tweets, which models both linguistic and psychological contexts.

Mishra et al.[66] utilized CNN to automatically extract cognitive features from the eye-movement (or gaze) data to enrich information for sarcasm detection. Word embeddings are also used for irony recognition in English tweets[120] and for controversial words identification in debates.[121]

## EMOTION ANALYSIS

Emotions are the subjective feelings and thoughts of human beings. The primary emotions include *love, joy, surprise, anger, sadness* and *fear*. The concept of emotion is closely related to sentiment. For example, the strength of a sentiment can be linked to the intensity of certain emotion like *joy* and *anger*. Thus, many deep learning models are also applied to emotion analysis following the way in sentiment analysis.

Wang et al.[122] built a bilingual attention network model for code-switched emotion prediction. A LSTM model is used to construct a document level representation of each post, and the attention mechanism is employed to capture the informative words from both the monolingual and bilingual contexts.

Zhou et al.[123] proposed an emotional chatting machine to model the emotion influence in large-scale conversation generation based on GRU. The technique has also been applied in other papers.[39,72,115]

Abdul-Mageed and Ungar[124] first built a large dataset for emotion detection automatically by using distant supervision and then used a GRU network for fine-grained emotion detection.

Felbo et al.[125] used millions of emoji occurrences in social media for pretraining neural models in order to learn better representations of emotional contexts.

A question-answering approach is proposed using a deep memory network for emotion cause extraction.[126] Emotion cause extraction aims to identify the reasons behind a certain emotion expressed in text.

**MULTIMODAL DATA FOR SENTIMENT ANALYSIS**

Multimodal data, such as the data carrying textual, visual, and acoustic information, has been used to help sentiment analysis as it provides additional sentiment signals to the traditional text features. Since deep learning models can map inputs to some latent space for feature representation, the inputs from multimodal data can also be projected simultaneously to learn multimodal data fusion, for example, by using feature concatenation, joint latent space, or other more sophisticated fusion approaches. There is now a growing trend of using multimodal data with deep learning techniques.

Poria et al.[127] proposed a way of extracting features from short texts based on the activation values of an inner layer of CNN. The main novelty of the paper is the use of a deep CNN to extract features from text and the use of multiple kernel learning (MKL) to classify heterogeneous multimodal fused feature vectors.

Bertero et al.[128] described a CNN model for emotion and sentiment recognition in acoustic data from interactive dialog systems.

Fung et al.[129] demonstrated a virtual interaction dialogue system that have incorporated sentiment, emotion and personality recognition capabilities trained by deep learning models.

Wang et al.[130] reported a CNN structured deep network, named **Deep Coupled Adjective and Noun** (DCAN) neural network, for visual sentiment classification. The key idea of DCAN is to harness the adjective and noun text descriptions, treating them as two (weak) supervision signals to learn two intermediate sentiment representations. Those learned representations are then concatenated and used for sentiment classification.

Yang et al.[131] developed two algorithms based on a conditional probability neural network to analyse visual sentiment in images.

Zhu et al.[132] proposed a unified CNN-RNN model for visual emotion recognition. The architecture leverages CNN with multiple layers to extract different levels of features (e.g., colour, texture, object, etc.) within a multi-task learning framework. And a bidirectional RNN is proposed to integrate the learned features from different layers in the CNN model.

You et al.[133] adopted the attention mechanism for visual sentiment analysis, which can jointly discover the relevant local image regions and build a sentiment classifier on top of these local regions.

Poria et al.[134] proposed some a deep learning model for multi-modal sentiment analysis and emotion recognition on video data. Particularly, a LSTM-based model is proposed for utterance-level sentiment analysis, which can capture contextual information from their surroundings in the same video.

Tripathi et al.[135] used deep and CNN-based models for emotion classification on a multimodal dataset DEAP, which contains electroencephalogram and peripheral physiological and video signals.

Zadeh et al.[136] formulated the problem of multimodal sentiment analysis as modelling intra-modality and inter-modality dynamics and introduced a new neural model named Tensor Fusion Network to tackle it.

Long et al.[137] proposed an attention neural model trained with cognition grounded eye-tracking data for sentence-level sentiment classification. A **Cognition Based Attention** (CBA) layer is built for neural sentiment analysis.

Wang et al.[138] proposed a **Select-Additive Learning** (SAL) approach to tackle the confounding factor problem in multimodal sentiment analysis, which removes the individual specific latent representations learned by neural networks (e.g., CNN). To achieve it, two learning phases are involved, namely, a selection phase for confounding factor identification and a removal phase for confounding factor removal.

## RESOURCE-POOR LANGUAGE AND MULTILINGUAL SENTIMENT ANALYSIS

Recently, sentiment analysis in resource-poor languages (compared to English) has also achieved significant progress due to the use of deep learning models. Additionally, multilingual features also can help sentiment analysis just like multimodal data. In the same way, deep learning has been applied to the multilingual sentiment analysis setting.

Akhtar et al.[139] reported a CNN-based hybrid architecture for sentence and aspect level sentiment classification in a resource-poor language, Hindi.

Dahou et al.[140] used word embeddings and a CNN-based model for Arabic sentiment classification at the sentence level.

Singhal and Bhattacharyya[141] designed a solution for multilingual sentiment classification at review/sentence level and experimented with multiple languages, including Hindi, Marathi, Russian, Dutch, French, Spanish, Italian, German, and Portuguese. The authors applied machine translation tools to translate these languages into English and then used English word embeddings, polarities from a sentiment lexicon and a CNN model for classification.

Joshi et al.[142] introduced a sub-word level representation in a LSTM architecture for sentiment classification of Hindi-English code-mixed sentences.

## OTHER RELATED TASKS

There are also applications of deep learning in some other sentiment analysis related tasks.

**Sentiment Intersubjectivity:** Gui et al.[143] tackled the intersubjectivity problem in sentiment analysis, where the problem is to study the gap between the surface form of a language and the corresponding abstract concepts, and incorporate the modelling of intersubjectivity into a proposed CNN.

**Lexicon Expansion:** Wang et al.[144] proposed a PU learning-based neural approach for opinion lexicon expansion.

**Financial Volatility Prediction:** Rekabsaz et al.[145] made volatility predictions using financial disclosure sentiment with word embedding-based information retrieval models, where word embeddings are used in similar word set expansion.

**Opinion Recommendation**: Wang and Zhang[146] introduced the task of opinion recommendation, which aims to generate a customized review score of a product that the particular user is likely to give, as well as a customized review that the user would have written for the target product if the user had reviewed the product. A multiple-attention memory network was proposed to tackle the problem, which considers users' reviews, product's reviews, and users' neighbours (similar users).

**Stance Detection**: Augenstein et al.[147] proposed a bidirectional LSTMs with a conditional encoding mechanism for stance detection in political twitter data. Du et al.[148] designed a target-specific neural attention model for stance classification.

## CONCLUSION

Applying deep learning to sentiment analysis has become a popular research topic lately. In this paper, we introduced various deep learning architectures and their applications in sentiment analysis. Many of these deep learning techniques have shown state-of-the-art results for various sentiment analysis tasks. With the advances of deep learning research and applications, we believe that there will be more exciting research of deep learning for sentiment analysis in the near future.

### Acknowledgments

Bing Liu and Shuai Wang's work was supported in part by National Science Foundation (NSF) under grant no. IIS1407927 and IIS-1650900, and by Huawei Technologies Co. Ltd with a research gift.